\documentclass[suppldata]{interact}
\usepackage{fix-cm}
\usepackage{epstopdf}
\usepackage{graphicx}
\usepackage{float}
\usepackage{footnote}
\usepackage{xfrac}
\usepackage{here}
\usepackage{url}
\usepackage{subcaption} 
\usepackage[margin=10pt,font={sf,small},labelfont=sf]{caption}
\usepackage{color}
\usepackage{natbib}
\usepackage{algorithm}
\usepackage{algpseudocode}
\usepackage{etoolbox}

\theoremstyle{plain}

\theoremstyle{definition}

\theoremstyle{remark}

\newcommand{\be}{\begin{equation}}
\newcommand{\ee}{\end{equation}}

\makeatletter

\newcommand*{\algrule}[1][\algorithmicindent]{
  \makebox[#1][l]{
    \hspace*{.2em}
    \vrule height .75\baselineskip depth .25\baselineskip
  }
}

\newcount\ALG@printindent@tempcnta
\def\ALG@printindent{
    \ifnum \theALG@nested>0
    \ifx\ALG@text\ALG@x@notext
    \else
    \unskip
    \ALG@printindent@tempcnta=1
    \loop
    \algrule[\csname ALG@ind@\the\ALG@printindent@tempcnta\endcsname]
    \advance \ALG@printindent@tempcnta 1
    \ifnum \ALG@printindent@tempcnta<\numexpr\theALG@nested+1\relax
    \repeat
    \fi
    \fi
}
\patchcmd{\ALG@doentity}{\noindent\hskip\ALG@tlm}{\ALG@printindent}{}{\errmessage{failed to patch}}
\patchcmd{\ALG@doentity}{\item[]\nointerlineskip}{}{}{}
\makeatother

\begin{document}

\title{\begin{center} An AI Approach for Learning the Spectrum of the Laplace-Beltrami Operator \end{center}}

\author{
\begin{center}
\name{Yulin An\textsuperscript{$\ast$}\thanks{Y.A. e-mail: yba5115@psu.edu} and Enrique del Castillo\textsuperscript{$\ast,\dagger$}\thanks{E.D.C., corresponding author, e-mail: exd13@psu.edu}}
\textsuperscript{$\ast$} Department of Industrial and Manufacturing Engineering\\
The Pennsylvania State University, University Park, PA\\
 \textsuperscript{$\dagger$}Department of Statistics\\ The Pennsylvania State University, University Park, PA
\end{center}
}

\maketitle

\begin{abstract}
The spectrum of the Laplace-Beltrami (LB) operator is central in geometric deep learning tasks, capturing intrinsic properties of the shape of the object under consideration. The best established method for its estimation, from a triangulated mesh of the object, is based on the Finite Element Method (FEM), and computes the top $k$ LB eigenvalues with a complexity of $O(Nk)$, where $N$ is the number of points. This can render the FEM method inefficient when repeatedly applied to databases of CAD mechanical parts, or in quality control applications where part metrology is acquired as large meshes and decisions about the quality of each part are needed quickly and frequently. As a solution to this problem, we present a geometric deep learning framework to predict the LB spectrum efficiently given the CAD mesh of a part, achieving significant computational savings without sacrificing accuracy, demonstrating that the LB spectrum is learnable. The proposed Graph Neural Network architecture uses a rich set of part mesh features —including Gaussian curvature, mean curvature, and principal curvatures. In addition to our trained network, we make available, for repeatability, a large curated dataset of real-world mechanical CAD models derived from the publicly available ABC dataset used for training and testing. Experimental results show that our method reduces computation time of the LB spectrum by approximately 5 times over linear FEM while delivering competitive accuracy.
\end{abstract}


\section{Introduction}
\label{sec:introduction}
\noindent The Laplace-Beltrami (LB) operator, a cornerstone of Riemannian geometry, generalizes the Laplacian to functions defined on manifolds. Its spectrum encodes rich geometric information of the manifold \cite{reuter2006laplace}, making it essential for geometric deep learning and computer vision in many fields of importance for industry, such as Computer-Aided-Design (CAD) model retrieval \cite{lin20183d}, part inspection \cite{wang2024ieee}, and statistical process control \cite{zhao2021intrinsic}. In these applications,  3-dimensional (3D) mechanical parts are either measured or modeled by triangulated meshes describing the geometry of the part.  Estimating the LB spectrum of a given part requires discretizing the operator on the mesh and numerically solving the resulting eigenvalue problem. The Finite Element Method (FEM) provides a sparse and most accurate estimate of the top $k$ eigenvalues of the LB operator with a computational complexity of $O(kN)$ where $N$ is the number of points in the mesh. The computational time for large meshes may render the FEM estimator too slow for the aforementioned manufacturing applications, where decisions are needed quickly and the repeated computation of the LB spectrum of many different parts is necessary.

In this paper, we take a different view on the computation of the LB spectrum, showing that it is a learnable geometric feature, given enough (and curated) meshes-spectra pairs from a database of real mechanical CAD models. We show how the proposed deep geometric framework predicts the LB spectrum of a given new part five times faster on average than how fast FEM methods can compute it (and up to two orders of magnitude faster than linear FEM computational schemes if ran on a GPU), without significance loss of accuracy, making its use in repetitive applications in engineering design and manufacturing feasible. The following fundamental definitions are used in the sequence of the paper.

{\bf Basic definitions.-} Let $\mathcal{M} \in \mathbb{R}^n$ be a $k$-dimensional compact Riemannian submanifold embedded in $\mathbb{R}^n$, and $f \in \mathcal{L}^2(\mathcal{M})$ denote a square-integrable function defined on $\mathcal{M}$. The {Laplace-Beltrami (LB) operator} acting on a function $f$ on $\mathcal{M}$ is defined as the differential operator \cite{chavel1984eigenvalues,rosenberg1997laplacian}:
\begin{equation}
\label{LB operator}
\Delta_{\mathcal{M}} f = - \mbox{div}_{\mathcal{M}}(\nabla_{\mathcal{M}} f)
\end{equation}
i.e., it is the divergence of the gradient field of $f$, where the negative sign is just a convention. If $\mathcal{M}=\mathbb{R}^n$, the LB operator $\Delta_{\mathcal{M}}$ reduces to the ordinary Laplacian:
\begin{equation}
\label{Laplacian operator}
\Delta f(x) = - \sum_{i=1}^n \frac{\partial^2 f}{\partial x_i^2}(x)
\end{equation}
 As such, when evaluated at a particular point $\bf x \in \mathbb{R}^n$, the operator models the local curvature of $f$ at $\bf x$. For a function defined on a general manifold $\mathcal{M}$, $\Delta_{\mathcal{M}}f(x)$ contains curvature information of $f(x)$ \textit{and} of $\mathcal{M}$ itself. In this paper, we deal with the simple case of 3-dimensional objects, i.e., $k=2$ and $n=3$. In this case, an intuitive explanation of the action of the LB operator is obtained for a parametric surface (or 2-manifold) $p(u,v) = (x(u,v),y(u,v),z(u,v)), (u,v) \in D \subset \mathbb{R}^2$, where the following relation holds \cite{chavel1984eigenvalues}:
\begin{equation}
\label{Geometric_Interpretation}
\Delta_{\mathcal{M}}\mathbf{p}(u,v) = - \mbox{div}_{\mathcal{M}}(\nabla_{\mathcal{M}}\mathbf{p}(u,v)) = 2H\mathbf{n}(u,v),
\end{equation}
where $\mathbf{n}(u,v)$ is the normal vector at the point $\mathbf{p}(u,v)$ on the surface $\mathcal{M}$ and $H$ is the mean curvature of $\mathcal{M}$ at $\mathbf{p}(u,v)$. Expression (\ref{Geometric_Interpretation}) can be visualized as a vector field of normals on each point $\mathbf{p}$ on the surface of $\mathcal{M}$ such that their height is twice the mean curvature of $\mathcal{M}$ at that point. This expression will prove important in the sequence, in our discussion of the input features used for learning the spectrum.

The LB operator appears in partial differential equations (PDEs) governing diffusion processes, like the heat equation, and this leads to its eigendecomposition, and therefore, its spectrum, which is our main objective. Separating the spatial part of the heat equation yields the expression:
\begin{equation}
    \Delta_{\mathcal{M}} f(x) =  \lambda f(x)
    \label{Helmholtz}
\end{equation}
 called the Helmholtz partial differential equation,  where the collection of eigenvalues $\{\lambda_i\}_{i=0}^{\infty}$ make up the {\em spectrum} of the LB operator and $\{\phi_i\}_{i=0}^{\infty}$ are the associated LB {eigenfunctions}. For compact manifolds without boundary, the spectrum is discrete
\cite{chavel1984eigenvalues}. The eigenfunctions provide an orthogonal basis for functions on $\mathcal{M}$. Since the LB operator is a continuous operator, it has an infinite number of eigenvalue-eigenfunction pairs \cite{rosenberg1997laplacian}. In practice, however, estimating the LB spectrum typically requires discretizing the operator on a triangulated mesh and numerically solving the resulting eigenvalue problem. This leads to a matrix operator acting on finite vectors. The Finite Element Method (FEM) provides a sparse and most accurate eigendecomposition of the LB operator by approximating the so-called weak form (i.e. the
classical Galerkin variational formulation, see \cite{zhao2022registration}) of the Helmholtz equation (\ref{Helmholtz}). For a mesh with $N$ vertices representing a surface (2-manifold) ${\cal M} \subset \mathbb{R}^3$, finite element methods solve Helmholtz equation by constructing sparse stiffness $\mathbf{K}$ and mass $\mathbf{M}$ matrices, yielding the generalized eigenvalue problem:
\begin{equation}
    \mathbf{K} \mathbf{U} = \lambda \mathbf{M} \mathbf{U},
    \label{eq:fem_eigenproblem}
\end{equation}
where $\mathbf{U}$ represents discrete eigenfunction coefficients. Although assembling $\mathbf{K}$ and $\mathbf{M}$ scales as $O(N)$, solving for the first $k$ eigenvalues via iterative solvers such as the Arnoldi algorithm demands $O(N k)$ operations, with memory requirements of $O(N)$ \cite{reuter2006laplace, zhao2022registration}. 

\begin{figure}[ht]
\centerline{\includegraphics[width=\columnwidth]{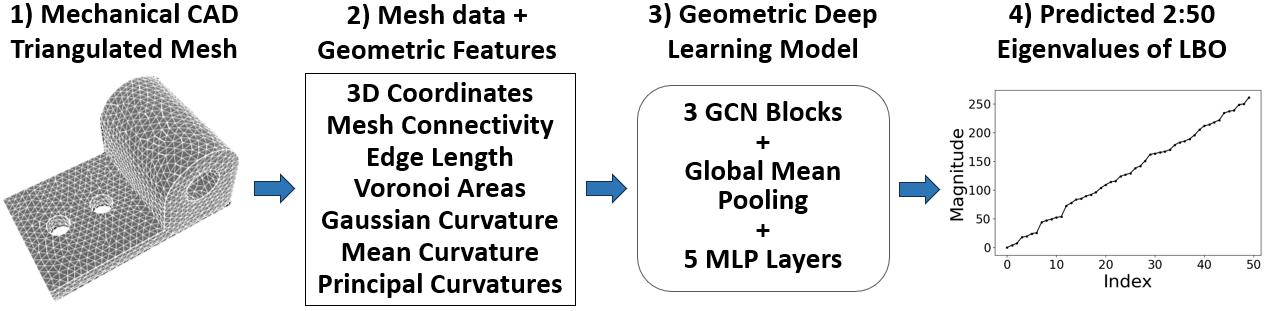}}
\caption{Schematic of the proposed framework: (1) input (triangulated mesh, typically a CAD model); (2) geometric feature extraction; (3) GCN architecture encoding features; (4) output  (2nd to 50th LB eigenvalues).}
\label{fig:pipeline}
\end{figure}

{\bf Overview of the proposed approach.-} We propose a geometric deep learning framework using graph convolutional networks (GCNs) to predict the top 50 LB eigenvalues directly from triangulated meshes, bypassing the need for eigenvalue solvers. We focus on single-component mechanical CAD models — prevalent in engineering design. Since the first LB eigenvalue is zero for single-component meshes, we define the learning task as finding a mapping $f: \mathcal{M} \to \mathbb{R}^{49}$, where $\mathcal{M}$ represents input meshes and the desired output is $[\lambda_2, \lambda_3, \ldots, \lambda_{50}]$. The goal is to accurately predict the spectrum while enhancing computational efficiency over FEM alternatives. We use a supervised learning approach, where mesh-spectrum pairs from a real-world mechanical CAD dataset are used for training, with the LB spectrum computed with a linear FEM estimator. The pipeline, shown in Figure~\ref{fig:pipeline}, extracts geometric features from the mesh, processes them via a GCN model, and outputs the predicted spectrum.

As it will be shown, the GCN framework achieves a speedup of 5$\times$ (CPU) to 100$\times$ (GPU) over linear FEM using the Python eigen-solver $\tt LaPy$ \cite{reuter2006laplace, wachinger2015brainprint} while maintaining competitive accuracy. Using Peak Signal-to-Noise Ratio (PSNR) as the evaluation metric (equation \ref{eq:PSNR} below), where predictions with $\mbox{PSNR} > 40$ are deemed accurate, our model attains $99.3\%$ performance on a curated dataset of more than 33,000 real-world mechanical CAD models derived from the ABC dataset \cite{Koch_2019_CVPR}.  Our primary contribution is a scalable learning-based solution to the LB-spectrum problem for large-scale CAD models, enabled by: (1) a curated dataset of real-world CAD meshes from the ABC dataset, offering a robust training foundation, and (2) an optimized GCN architecture tailored to encode geometric features for precise spectral prediction. This approach effectively addresses challenges in geometric variability, spectral precision, and mesh irregularities.

The rest of the paper is organized as follows. In the next section, we first review related work on operator learning and geometric deep learning (GDL) for CAD models. Section~\ref{sec:methodology} formalizes the prediction problem, details the geometric features used, and describes the GCN architecture; Section~\ref{sec:experiments} presents training and evaluation on the dataset, a curated version of the ABC CAD model database \cite{Koch_2019_CVPR}. The paper ends with a summary of findings, limitations, and implications for further research. All Python code, trained GCN, and curated dataset of CAD parts are made available as supplementary materials.

\section{Related Work}
\label{sec:related}
\noindent The advent of deep learning has spurred interest in learning-based approximations of operators occurring in Partial Differential Equations (PDEs). Pioneered by \cite{chen1995universal}, this paradigm employs deep learning to develop efficient surrogate models for many-query tasks, complementing traditional numerical methods. Unlike conventional neural networks that output scalars or fixed vectors, operator learning finds a map between infinite-dimensional function spaces, yielding functional outputs. Two main approaches prevail: earlier encoder-decoder architectures, such as PCA-Net and DeepONet, and recent neural operator frameworks, including the Fourier Neural Operator, Graph Neural Operator, Convolutional Neural Operator, and the Spectral Neural Operator (for a review of these architectures, see \cite{kovachki2024operator}).

Deep learning (DL) for 3D CAD models and point clouds has primarily focused on classification and segmentation \cite{yi2017SyncSpecCNN}, with limited exploration of differential operator learning on such data. While convolutional neural networks (CNNs) dominate 2D image analysis, including RGB-D data from LiDAR \cite{hua2018pointwise}, few architectures target 3D meshes or point clouds. For instance, \cite{zhang2020view} proposed a view-based 3D CAD model retrieval method using deep residual networks (ResNets), but its dependence on multiview representations (solid and line views) limits robustness. In \cite{wang2019learning}, the authors introduced a CNN to learn geometric and PDE operators from mesh data via a parametric family of linear operators, potentially including the LB operator; however, the lack of implementation details and follow-up studies restricts its impact. Similarly, \cite{smirnov2022deep} developed HodgeNet to learn operators from mesh data, but the extrinsic nature of the Hodge Laplacian and its reliance on explicit eigendecomposition, rather than a learned spectrum, limit its applicability to tasks such as LB eigenvalue prediction. 

Geometric deep learning, as surveyed in \cite{bronstein2017geometric}, extends DL to non-Euclidean domains, providing a foundation for our graph convolutional network (GCN)-based approach. A recent paper on GNCs related to our work is \cite{wu2024graph}, which first addressed the LB eigenvalue prediction problem. However, their results were constrained to predicting only the first ten LB eigenvalues of synthetic parts in a small dataset of ten simple 3D geometry classes, with neither code nor a complete dataset made available. Focusing on so few eigenvalues makes it impossible to model even moderately complex mechanical parts.  As it is well-known, higher LB eigenvalues and their eigenvectors capture increasingly finer geometrical details of a mesh, in a manner analogous to how higher harmonics in Fourier analysis approximate finer features of functions defined on a circle \cite{rosenberg1997laplacian}. Thus, the more LB eigenvalues are accurately predicted, the more usability of the deep geometric model architecture becomes, making it an increasingly more useful solution for industrial applications, given that it applies to parts with more complex geometrical details. One of the key differences with our DL approach is that it predicts the first 50 LB eigenvalues, and hence has applicability to handle a wide variety of parts of considerably intricate geometry, as demonstrated below. Our architecture is tailored to predicting these many eigenvalues, and we have experimented, as will be shown below, with many different configurations and loss functions. A second key difference is that we approach the LB spectrum prediction problem based on a large dataset (the ``ABC" dataset, \cite{Koch_2019_CVPR}), comprising 100,000 high-quality CAD meshes, to train and evaluate the proposed architecture on diverse, real-world mechanical part geometries. We curated this database, providing over 33,000 real-world single-component mechanical CAD models.  Finally, our codes, trained deep network, and curated dataset are made available to other researchers as supplementary materials (see Appendix \ref{app: supplementary}).

\section{Methodology}
\label{sec:methodology}

\subsection{Problem Definition}
\noindent Given a dataset of mechanical CAD models, this work aims to learn a mapping $f: \mathcal{M} \to \mathbb{R}^{49}$, where $\mathcal{M}$ denotes the space of mechanical CAD meshes, and the output comprises the 2nd to 50th Laplace-Beltrami (LB) eigenvalues. We focus on models with a single connected component, discarding the first eigenvalue, which is known to always be zero for connected objects \cite{chavel1984eigenvalues}. Thus, our GCN approach targets the 2nd through the 50th LB eigenvalues of a compact, smooth Riemannian manifold embedded in 3D Euclidean space, represented as a triangulated surface mesh.

\subsection{Geometric features used for training}
\begin{figure}[ht]  
    \centering  
    \includegraphics[scale=0.5]{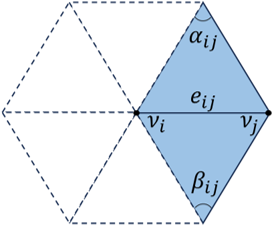}  
    \caption{Illustration of a first-order neighbor of a vertex, $v_i$, and two angles, $\alpha_{ij}$ and $\beta_{ij}$, opposite to an edge $e_{ij}$.}  
    \label{fig: Cotangent weights}  
\end{figure}

\noindent We extract a comprehensive set of geometric features from each triangular mesh $\mathcal{M}$ and provide them as inputs to the GCN. Each mesh is modeled as an undirected graph $G = (V, E, F)$ where $V = \{v_1, v_2, \dots, v_N\}$ is the set of vertices (mesh points), $E = \{e_{ij} \mid v_i, v_j \in V\}$ is the set of edges connecting two vertices, and $F = \{f_k \mid k = 1, \dots, M\}$ is the set of all triangular faces. Vertices are parameterized by their 3D Cartesian coordinates $v_i = (x_i, y_i, z_i), \quad i = 1, \dots, N$, where $N$ is the total vertex count, and provide the spatial basis for geometric analysis. Edges $e_{ij}$ connect vertex pairs $v_i, v_j$ and encode local connectivity essential for computation of {\em differential} properties (recall the LB operator is a differential operator). In addition to these baseline mesh elements, we introduce normalized edge lengths as edge features, alongside vertex features comprising mixed Voronoi areas, unweighted discrete Gaussian curvatures, unweighted discrete mean curvatures, and principal curvatures. This set of features was selected to balance local and global geometric cues, improving spectral prediction over using only baseline coordinates and connectivity mesh data. These features, explained below, collectively capture intrinsic and extrinsic geometric properties critical to LB eigenvalue prediction, and they form the input to our GCN architecture, detailed in the next subsection. We note how an underlying goal when using the following geometrical features was to make them as independent of each other as possible, to avoid redundancies that would not promote learning.
\\
\\
\textbf{Normalized Edge Length.-}  
The length of an edge $e_{ij}$ connecting two neighboring vertices $v_i$ and $v_j$ is given by:
\begin{equation}
l_{ij} = \sqrt{(x_i - x_j)^2 + (y_i - y_j)^2 + (z_i - z_j)^2}
\end{equation}
We standardize the scale of edge lengths across different meshes, normalizing $l_{ij}$ by the maximum edge length over the entire mesh \cite{wu2024graph}:
\begin{equation}
\tilde{l}_{ij} = \frac{l_{ij}}{\underset{e_{ij} \in E}{\max} l_{ij}}
\end{equation}
\\
\textbf{Mixed Voronoi Areas.-}  
The mixed Voronoi area \cite{meyer2003discrete} associated with a vertex $v_i$ provides an estimate of the local surface area contribution of that vertex. It is computed as:
\begin{equation}
A_i = \frac{1}{8} \sum_{j \in \mathcal{N}(i)} (cot \alpha_{ij} + cot \beta_{ij}) \|v_i - v_j \|^2,
\end{equation}
where $\mathcal{N}(i)$ represents the set of neighboring vertices of $v_i$, and $A_{ij}$ is the area of the triangle formed by $v_i$ and its adjacent neighbors. We compute the mixed Voronoi cell areas
with the $\tt libigl$ library \cite{libigl}.
\\
\\
\textbf{Unweighted Discrete Gaussian Curvatures.-}  
The Gaussian curvature quantifies the intrinsic curvature of a surface at a given point. For a mesh vertex $v_i$, the discrete Gaussian curvature is traditionally computed using the angle deficit formula normalized by the Voronoi area:  
\begin{equation}  
K_i = \frac{1}{A_i} \left( 2\pi - \sum_{\theta_j \in \mathcal{N}(i)} \theta_j \right),
\end{equation}  
where $ \theta_j $ are the interior angles of the triangles incident on $ v_i $, and $ A_i $ is the mixed Voronoi area. However, as we wish to incorporate the $A_i$'s as an independent input feature, we use instead an {\em unweighted} discrete Gaussian curvature formulation:  
\begin{equation}  
\tilde{K}_i = 2\pi - \sum_{\theta_j \in \mathcal{N}(i)} \theta_j.
\end{equation}  
We utilize the $\tt igl$ Python library \cite{libigl} to compute the unweighted discrete Gaussian curvature.  
\\
\\
\textbf{Unweighted Discrete Mean Curvatures.-}  
Although the LB eigenvalues encode global geometric properties of an object, the relation between the LB operator and the mean curvature given by equation  (\ref{Geometric_Interpretation}) indicates that this type of curvature is an important element to consider for training. The mean curvature measures the local bending of a surface, and for a mesh vertex $v_i$, the discrete mean curvature \cite{mari2019geometric} is computed as:  
\begin{equation}  
H_i = \frac{1}{2A_i} \sum_{j \in \mathcal{N}(i)} (\cot \alpha_{ij} + \cot \beta_{ij}) \|v_i - v_j \|,
\end{equation}  
where $ \alpha_{ij} $ and $ \beta_{ij} $ are the angles opposite to edge $ e_{ij} $ in the two adjacent triangles, as illustrated in Figure \ref{fig: Cotangent weights}.
\\
\\
\noindent Following the same rationale as before, we use instead an {\em unweighted} discrete mean curvature:  
\begin{equation}  
\tilde{H}_i = \sum_{j \in \mathcal{N}(i)} (\cot \alpha_{ij} + \cot \beta_{ij}) \|v_i - v_j \|.
\end{equation}  
\\
\\
\textbf{Principal Curvatures.-}  
Principal curvatures, $\kappa_1$ and $\kappa_2$, represent the maximum and minimum normal curvatures at a given vertex. They are related to the Gaussian and mean curvatures:
\begin{equation}
\kappa_1, \kappa_2 = H \pm \sqrt{H^2 - K}.
\end{equation}
Given this relation and in order to avoid feeding redundant features, we compute them directly via quadratic fitting \cite{panozzo2010efficient} rather than the expression above. This method fits a quadric function to the local neighborhood of each vertex, analytically deriving $\kappa_1$ and $\kappa_2$, implemented using the \texttt{libigl}  Python library \cite{libigl}. 

\subsection{Justifying the Geometric Features Used for Training}
This section shows some preliminary training results in deciding the list of geometric features discussed in the previous section using a synthetic dataset created following the description by Wu et. al. \cite{wu2024graph}. 5000 parts are fabricated with random scales and orientations for each of the ten classes of objects and only the first 10 LB eigenvalues are being learned. All models were trained for 500 epochs using the SGD optimizer with a learning rate of $10^{-4}$, weight decay of $10^{-5}$, batch size of 16, and MLP layer widths of 256, 128, 64, 32, and 10. Table \ref{tab:prelim_performance} shows the training losses and performance metrics with various combinations of geometric features. This preliminary analysis confirms that, in addition to the features mentioned in Wu et. al. \cite{wu2024graph}, adding the unweighted curvature information further improves the prediction accuracy. We point out that the mesh information and some measure of curvature are obligatory, and different unweighted curvature features are used to make the features as independent of each other as possible, while a comprehensive study to validate the necessity of each remaining feature is reserved for future research. 

\begin{table}[ht]
    \centering
    \caption{Training of the GCN using different combinations of geometric features under $L_1$ loss function and $PSNR > 20$ evaluation metric. The model trained with the $L_1$ loss and different combinations of geometric features on the left column was evaluated over the test parts for $L_1$, $L_2$, Polar, and Relative Percentage Difference (RPD, see Section 4.2) losses in addition to the Peak Signal-to-Noise Ratio (PSNR, see Section 4.3) statistic. All GCNs have widths of 256, 128, 64, 32, and 10 channels in the 5-layer MLP. Bold numbers are the best values.}
    \label{tab:prelim_performance}
    \begin{tabular}{|l|c|c|c|c|c|c|}
        \hline
        \textbf{Model(Features)} & $\mathbf{L_1}$  & $\mathbf{L_2}$  & Polar & RPD & \textbf{\mbox{PSNR} $>$ 20} \\
        \hline
        $\text{GCN}$(Wu et. al.) & 1.2147 & 0.4879 & 0.0337 & 1.2868 & 0.9088  \\
        \hline
        $\text{GCN}({All_{weighted}})$ & 1.1652 & 0.4650 & 0.0313 & 1.2700 & 0.9196  \\
        \hline
        $\text{GCN}(All_{unweighted})$ & $\mathbf{0.4931}$ & $\mathbf{0.2057}$ & $\mathbf{0.0119}$ & $\mathbf{1.1137}$ & $\mathbf{0.9956}$  \\
        \hline
    \end{tabular}
\end{table}

\subsection{Model Architecture}
\begin{figure}[ht]
    \centering
    \includegraphics[width=\columnwidth]{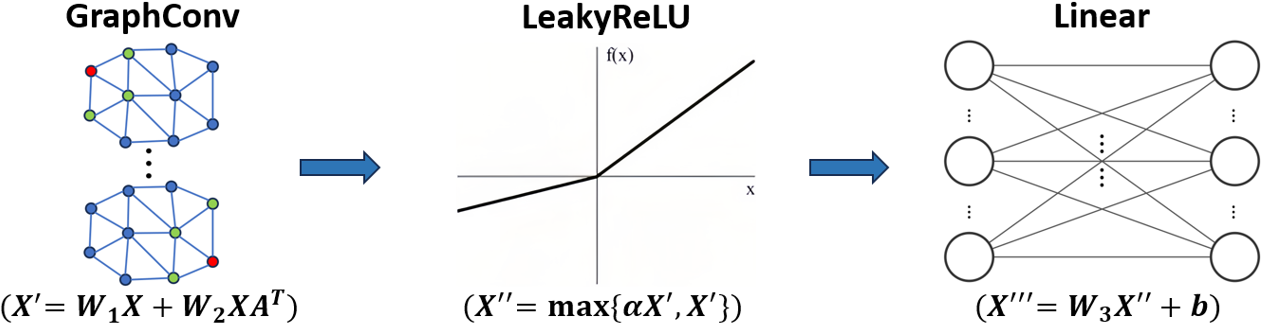}  
    \caption{Structure of a GCN block. A single GCN block is composed by 1) a Graph Convolutional layer, 2) a LeakyReLU activation function, and 3) a Linear layer.}
    \label{fig:GCN Block}
\end{figure}

\begin{figure*}[t]
    \centering
    \includegraphics[width=\columnwidth]{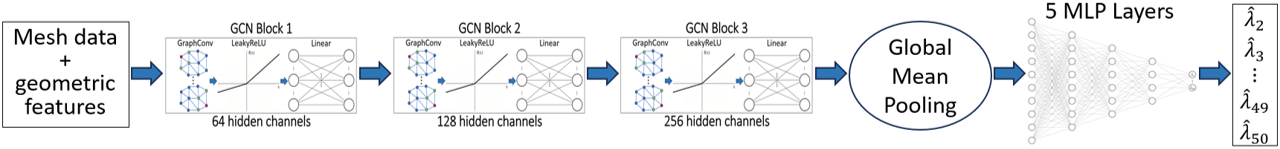}  
    \caption{GCN architecture for learning the 2nd to 50th eigenvalues of the Laplace-Beltrami operator using mesh data and geometric features. Input features are processed through 3 GCN blocks with 64, 128, and 256 hidden channels, respectively, followed by a global mean pooling layer, which is then followed by a 5-layer MLP with 8192, 4096, 2048, 1024, and 49 channels, respectively.}
    \label{fig:GDL_Architecture}
\end{figure*}

\noindent  The graph convolutional kernel in our architecture follows the formulation by~\cite{morris2021weisfeilerlemanneuralhigherorder}, implemented as \texttt{conv.GraphConv} in the \texttt{PyTorch\_Geometric} Python library:
\begin{equation}
    \mathbf{x}'_i = \mathbf{W}_1 \mathbf{x}_i + \mathbf{W}_2 \sum_{j \in \mathcal{N}(i)} e_{j,i} \mathbf{x}_j,
\end{equation}
where $\mathbf{W}_1$ and $\mathbf{W}_2$ are weight matrices of the graph convolutional kernel, $\mathcal{N}(i)$ is the first-order neighborhood of vertex $ v_i $, and $ e_{j,i} $ denotes the edge weight from source node $ v_j $ to target node $ v_i $. We use the normalized edge length $\tilde{l}_{ij}$ as the edge weight. A GCN block consists of a graph convolutional layer, a LeakyReLU activation function, and a linear layer, as illustrated in Figure~\ref{fig:GCN Block}.
\\
\\
The proposed architecture builds upon the framework introduced by \cite{wu2024graph}. The mean pooling layer was replaced with a global mean pooling layer to effectively aggregate graph-level features, enhancing robustness across diverse mesh structures. The width of the five multilayer perceptron (MLP) layers was increased, with widths of 8192, 4096, 2048, 1024, and 49, respectively. LeakyReLU activation functions, with a slope of 0.01, are applied consistently across both GCN blocks and MLP layers. Inspired by \cite{lee2019wide} and \cite{belkin2021fit}, which demonstrate the superior performance of wider networks, the progressively decreasing MLP widths balance computational efficiency with feature refinement, aligning with the dimensionality of the target eigenvalues. The final MLP layer outputs 49 channels, corresponding to the 2nd to 50th eigenvalues of the Laplace-Beltrami operator. This configuration yields a $60\%$ improvement in eigenvalue prediction accuracy compared to the narrower MLP designs, as detailed in section IV D. The proposed GCN architecture, illustrated in Figure \ref{fig:GDL_Architecture}, is implemented using the $\tt PyTorch$ and $\tt PyTorch\_Geometric$ libraries in Python, facilitating scalable mesh processing for industrial applications.

\section{Experiments}
\label{sec:experiments}

\subsection{Dataset}
\noindent  The dataset in this study is derived from the ABC normal estimation benchmark dataset \cite{Koch_2019_CVPR}, a subset of 100,000 predominantly mechanical CAD models extracted from the broader ABC Dataset, which comprises one million such models selected for geometric deep learning research sources \cite{Koch_2019_CVPR}. We first filtered this dataset to include only CAD models with a single connected component with no boundaries and a genus value less than three (i.e., those with 2 or fewer holes), using the $\tt pymeshlab$  Python library \cite{pymeshlab2021}. This process yielded a refined subset of 33,327 models.

To ensure numerical stability and improve the accuracy of LB spectrum estimation via the linear FEM \cite{an2025practical}, each mesh was pre-processed with the explicit isotropic remeshing algorithm (IRA) from $\tt pymeshlab$ \cite{pymeshlab2021}. This procedure enhances isotropy—distributing vertices more uniformly across the mesh surface—while targeting a vertex count between 1,750 and 2,250. However, 87 meshes were excluded due to remeshing failures caused by complex geometries and excessive geometric distortion. Additionally, we removed 397 duplicate CAD models—defined as geometrically identical and same in scale—along with two models exhibiting numerical anomalies, probably due to defects in the meshes, yielding a final dataset of 32,841 meshes. We deemed two CAD models $A$ and $B$ identical if the maximum difference between their first 50 LB eigenvalues satisfies $\underset{i=1,\ldots,50}{max} |\lambda_i^A - \lambda_i^B| < 10^{-8}$. The final curated dataset from the ABC normal estimation benchmark dataset is available as supplementary materials to this paper (see Appendix \ref{app: supplementary}).

The LB eigenvalues were computed using the $\tt LaPy$ Python library \cite{reuter2006laplace, wachinger2015brainprint}, with the 50th eigenvalue spanning a broad range ($5.8985 \times 10^{-2}$ to $1.4265 \times 10^{9}$), highlighting the dataset's inherent scale variability. To mitigate gradient instability during training, each mesh was normalized to fit within a unit cube and centered at the origin, resulting in a narrower range of $4.2326 \times 10^{2}$ to $4.9574 \times 10^{5}$. To recover approximate eigenvalue values in the original scale, predicted LB eigenvalues must be post-multiplied by the scale factor applied during normalization, given the known relation between the scale of the manifold and its LB spectrum \cite{reuter2006laplace,zhao2021intrinsic}. To enhance the model's ability to capture the rotational invariance of the LB operator, five random rotations about the origin were applied per mesh, expanding the dataset to 164,205 samples. Mesh manipulations, including translation, rotation, and scaling, were performed using the $\tt trimesh$ \cite{trimesh2019} Python library. 

\subsection{Loss Function}
\noindent We used the \textbf{Relative Percentage Difference, (RPD)} (also called the {\em Symmetric Mean Absolute Percentage Error} in the forecasting literature, see \cite{flores1986}), a widely used error or loss function in forecasting and quality control that is scale-invariant and quantifies the relative deviation between predicted and ground truth values. Given a vector of target values $y_i$'s and its corresponding prediction $\hat{y}_i $'s, the RPD loss is defined as:
\begin{equation}
\mathcal{L}_{\text{rpd}} = \sum_{i=2}^{50}\frac{| y_i - \hat{y}_i |}{|y_i| + |\hat{y}_i| + \epsilon},
\label{RPD}
\end{equation}  
where $\epsilon$ is a small constant to prevent division by zero. This formulation ensures that errors are evaluated in a manner robust to the absolute magnitude of the target variable, making it particularly effective for tasks where relative accuracy is more meaningful than absolute differences. Unlike standard mean squared error (MSE), $\text{L}_1$, or $\text{L}_2$ losses, the RPD loss naturally adapts to varying scales of data, mitigating issues related to dynamic range discrepancies. As a result, the RPD loss is particularly well-suited for evaluating the predictions of the LB spectrum \cite{wu2024graph}, which is a non-decreasing sequence that can exhibit a broad range of numbers.

\subsection{Evaluation of different models}
\begin{figure}[H]
    \centering
    \begin{subfigure}{0.20\textwidth}
        \includegraphics[width=\textwidth,height = 3.5cm]{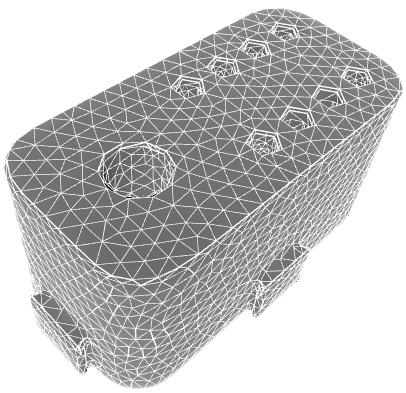}
    \end{subfigure}
    \begin{subfigure}{0.25\textwidth}
        \includegraphics[width=\textwidth,height = 3.5cm]{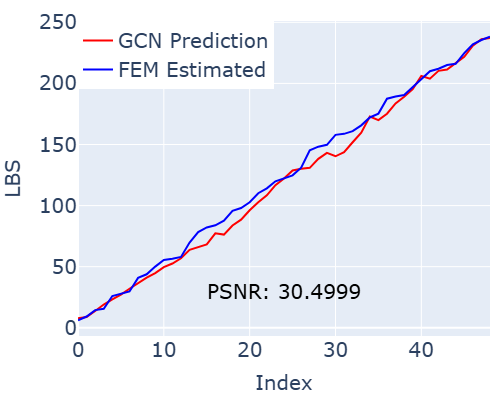}
    \end{subfigure}
    \hspace{0.5cm}
    \begin{subfigure}{0.20\textwidth}
        \includegraphics[width=\textwidth,height = 3.5cm]{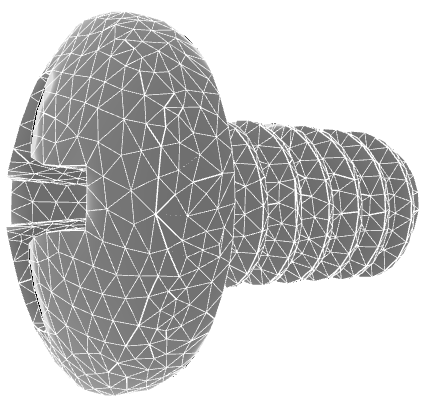}
    \end{subfigure}
    \begin{subfigure}{0.25\textwidth}
        \includegraphics[width=\textwidth,height = 3.5cm]{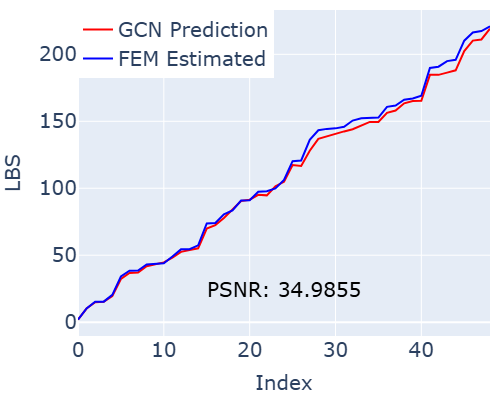}
    \end{subfigure}
    \vfill
    \begin{subfigure}{0.20\textwidth}
        \includegraphics[width=\textwidth,height = 3.5cm]{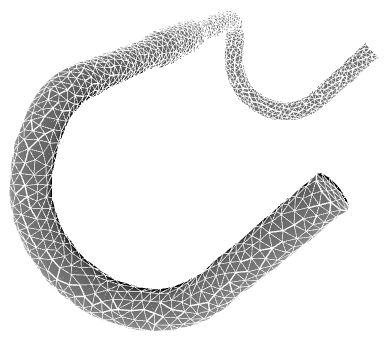}
    \end{subfigure}
    \begin{subfigure}{0.25\textwidth}
        \includegraphics[width=\textwidth,height = 3.5cm]{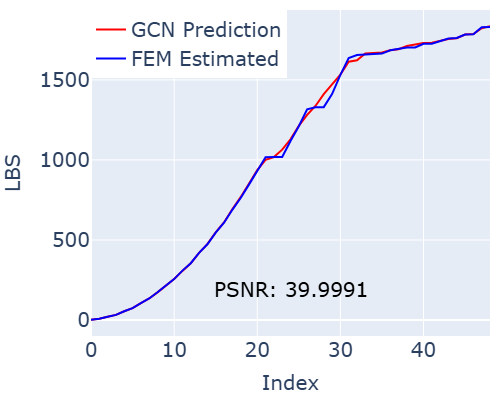}
    \end{subfigure}
    \hspace{0.5cm}
    \begin{subfigure}{0.20\textwidth}
        \includegraphics[width=\textwidth,height = 3.5cm]{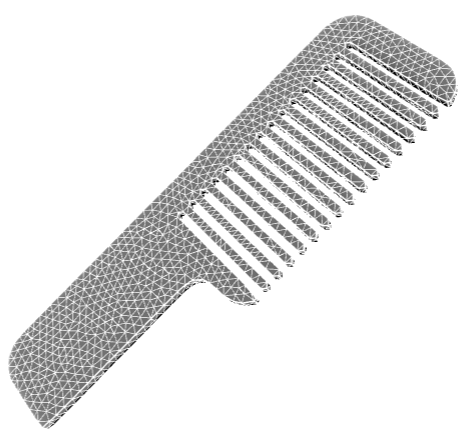}
    \end{subfigure}
    \begin{subfigure}{0.25\textwidth}
        \includegraphics[width=\textwidth,height = 3.5cm]{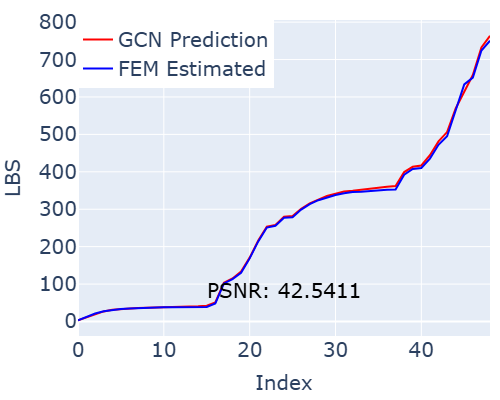}
    \end{subfigure}
    \vfill
    \begin{subfigure}{0.20\textwidth}
        \includegraphics[width=\textwidth,height = 3.5cm]{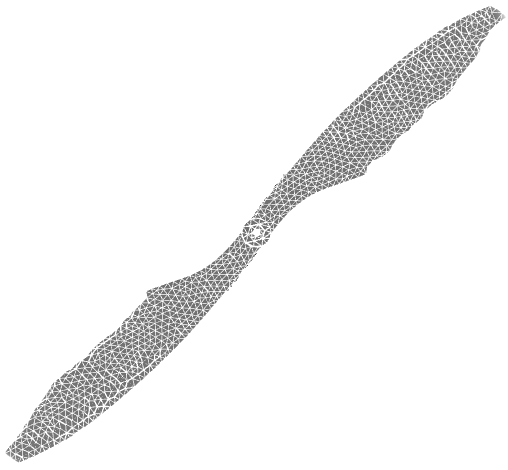}
    \end{subfigure}
    \begin{subfigure}{0.25\textwidth}
        \includegraphics[width=\textwidth,height = 3.5cm]{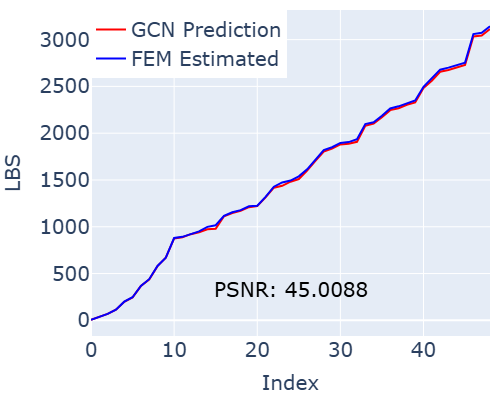}
    \end{subfigure}
    \hspace{0.5cm}
    \begin{subfigure}{0.20\textwidth}
        \includegraphics[width=\textwidth,height = 3.5cm]{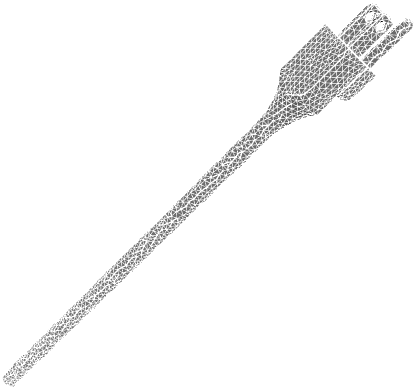}
    \end{subfigure}
    \begin{subfigure}{0.25\textwidth}
        \includegraphics[width=\textwidth,height = 3.5cm]{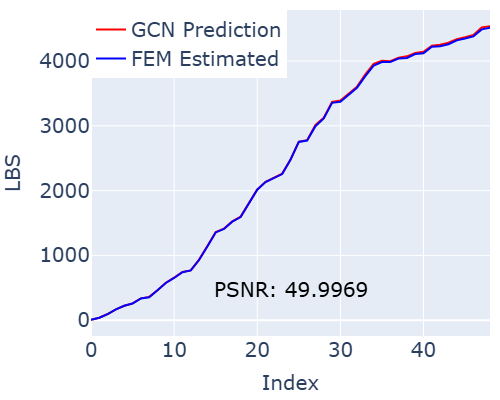}
    \end{subfigure}
    \vfill
    \begin{subfigure}{0.20\textwidth}
        \includegraphics[width=\textwidth,height = 3.5cm]{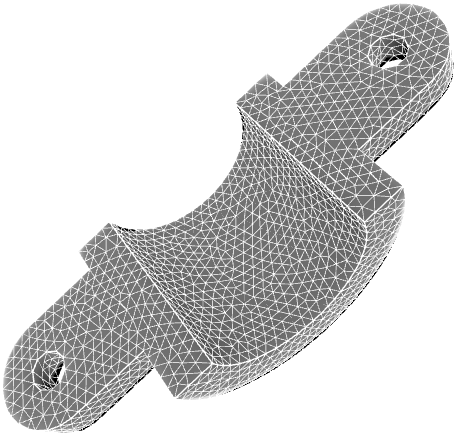}
    \end{subfigure}
    \begin{subfigure}{0.25\textwidth}
        \includegraphics[width=\textwidth,height = 3.5cm]{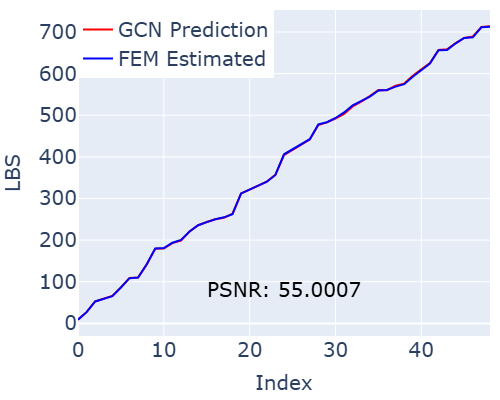}
    \end{subfigure}
    \hspace{0.5cm}
    \begin{subfigure}{0.20\textwidth}
        \includegraphics[width=\textwidth,height = 3.5cm]{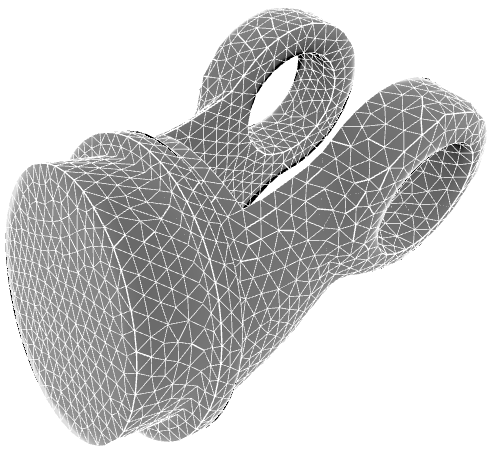}
    \end{subfigure}
    \begin{subfigure}{0.25\textwidth}
        \includegraphics[width=\textwidth,height = 3.5cm]{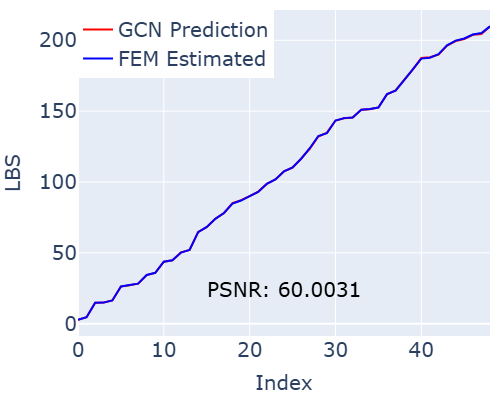}
    \end{subfigure}
    \vfill
    \begin{subfigure}{0.20\textwidth}
        \includegraphics[width=\textwidth,height = 3.5cm]{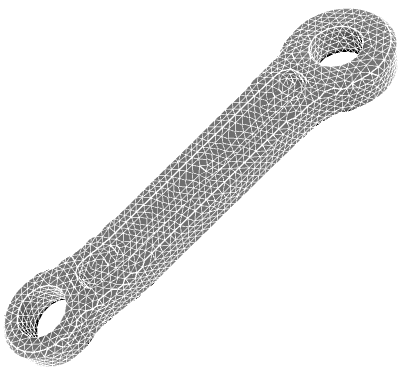}
    \end{subfigure}
    \begin{subfigure}{0.25\textwidth}
        \includegraphics[width=\textwidth,height = 3.5cm]{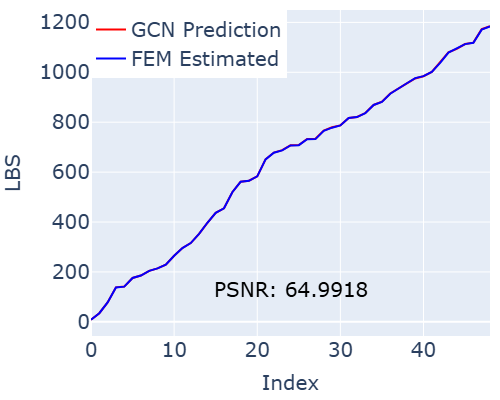}
    \end{subfigure}
    \hspace{0.5cm}
    \begin{subfigure}{0.20\textwidth}
        \includegraphics[width=\textwidth,height = 3.5cm]{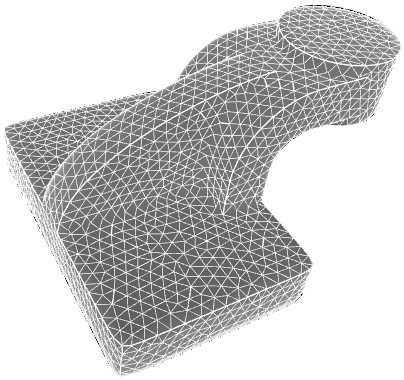}
    \end{subfigure}
    \begin{subfigure}{0.25\textwidth}
        \includegraphics[width=\textwidth,height = 3.5cm]{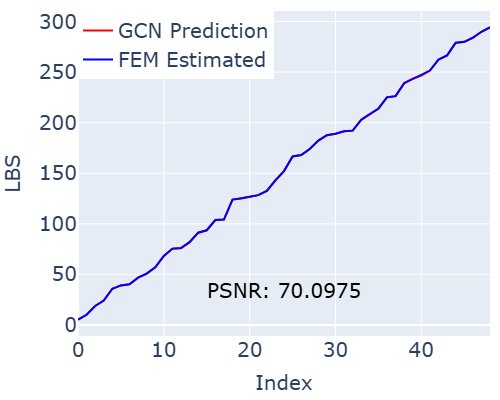}
    \end{subfigure}
    \caption{Sample predictions of the 2nd to 50th Laplace-Beltrami (LB) eigenvalues and their corresponding PSNR value. The predicted eigenvalue sequence (red) approaches the linear FEM ground truth (blue) as the PSNR value increases. Predictions were deemed accurate when $\mbox{PSNR} > 40$ (last 7 parts).}
    \label{fig:Sample_Output}
\end{figure}

To evaluate the different GCN configurations (some of which are detailed below), we used the \textbf{Peak Signal-to-Noise Ratio (PSNR)}, a widely used perceptual quality metric in image analysis for assessing the fidelity of reconstructed signals \cite{Korhonen2012}. Given a reference signal $ y $ and its corresponding reconstruction $ \hat{y} $, PSNR is defined as  
\begin{equation}
\mbox{PSNR($\mathbf{y},\mathbf{\hat{y}}$)} = 10 \log_{10} \left( \frac{(\max(\mathbf{y})-\min(\mathbf{y}))^2}{\text{MSE}(\mathbf{y}, \mathbf{\hat{y}})} \right),
\label{eq:PSNR}
\end{equation}  
where MSE$(\mathbf{y}, \mathbf{\hat{y}})$ represents the mean squared error between the original and reconstructed signals. Higher PSNR values indicate better reconstruction fidelity, as they correspond to lower distortion relative to the maximum possible signal intensity \cite{wu2024graph}. The PSNR is particularly valuable for benchmarking generative models, as it provides an interpretable, signal-level assessment of reconstruction accuracy. In our experiments, we found that comparing FEM and predicted spectra, if $\mbox{PSNR} > 40$ the prediction is very accurate and closely matches the ground truth. Some example parts, with their LB spectrum predictions and PSNR values, are shown in Figure \ref{fig:Sample_Output}.
 The model performance is reported as the percentage of parts in the testing dataset for which $\mbox{PSNR}>40$ is true. Note that this metric was not the loss function used in training; the loss used was RPD (eq. \ref{RPD}). We used PSRN only to assess the quality of the predictions in a more interpretable way and to select the best final model. 

\subsection{Training}
\noindent The dataset was randomly partitioned into training, validation, and testing sets (80:10:10 split), with a fixed seed ensuring reproducibility. All models were trained for 1000 epochs with the ADAM optimizer and batch size 16 on an NVIDIA A100 GPU with 40GB of memory and an Intel Xeon Gold 6248R CPU with 64 GB of memory. The best model was selected based on the highest performance using the $\mbox{PSNR} > 40$ evaluation metric. The final model, optimized with the RPD loss function, underwent a two-stage training process: first, 500 epochs with a learning rate of $10^{-4}$, weight decay of $10^{-5}$, and batch size of 16; then, an additional 500 epochs were run with a reduced learning rate of $10^{-5}$ (same weight decay and batch size). We next describe the different experiments that led to this final architecture and choice of loss function.

\subsection{Experimental tests to find the best possible GCN}
\label{sec:results}
\noindent The final GCN architecture, shown in figure \ref{fig:GDL_Architecture}, was found by iterative improvement over many experiments. Here we report some of those experiments and discuss their results.

{\em Selecting the best loss function for training.-}
To identify the most suitable loss function for training our model, we evaluated both convergence speed and stability across various loss functions. All models were trained for 1000 epochs using the ADAM optimizer with a learning rate of $10^{-4}$, weight decay of $10^{-5}$, batch size of 16, and MLP layer widths of 1024, 512, 256, 128, and 49. Table~\ref{tab:performance_of_various_loss_functions} shows the training losses and performance metrics for each model. We compared standard loss functions, including $L_1$ and $L_2$ losses, RPD loss, and the Polar loss proposed by \cite{wu2024graph}. The RPD loss demonstrated superior performance in both convergence speed and training stability, attributed to its robustness against the significant variation in the scale of the Laplace-Beltrami (LB) eigenvalues, given that it is a non-decreasing sequence. Training each model with a different loss function required approximately 44 hours on an NVIDIA A100 GPU (40 GB memory) paired with an Intel Xeon Gold 6248R CPU (64 GB memory). Notably, RPD training provided not only the lowest RPD loss {\em but also} the lowest Polar loss. Hence, the use of the Polar loss, which in addition provides very poor PSNR performance, is unjustified.
\begin{table}[ht]
    \centering
    \caption{Training of the GCN using different loss functions:  $L_1$, $L_2$, Polar, and RPD. The model trained with the loss on the left column was evaluated over the test parts for all four losses in addition to the PSNR statistic. All GCNs have widths of 1024, 512, 256, 128, and 49 channels in the 5-layer MLP. Bold numbers are best per loss function.}
    \label{tab:performance_of_various_loss_functions}
    \begin{tabular}{|l|c|c|c|c|c|}
        \hline
        \textbf{Model(loss)} & $\mathbf{L_1}$ & $\mathbf{L_2}$ & \textbf{Polar} & \textbf{RPD} & \textbf{\mbox{PSNR} $>$ 40} \\
        \hline
        $\text{GCN}(L_1)$ & \textbf{260.74} & 49.31 & 0.0116 & 0.8785 & 0.2169  \\
        \hline
        $\text{GCN}({L_2})$ & 262.71 & \textbf{47.63} & 0.0121 & 0.9709 & 0.2006  \\
        \hline
        $\text{GCN}(Polar)$ & 852.02 & 155.95 & 0.0222 & 1.5586 & 0.0366  \\
        \hline
        $\text{GCN}({RPD})$ & 284.82 & 57.35 & \textbf{0.0106} & \textbf{0.4984} & \textbf{0.2764}  \\
        \hline
    \end{tabular}
\end{table}

{\em Selecting the best width of the MLP layers.-}
To determine the optimal number of hidden channels in the MLP layers, we tried to balance prediction time and prediction performance. We evaluated four GCN models with 1024, 2048, 4096, and 8192 hidden channels in the first MLP layer. Subsequent MLP layers have half the channels of the preceding layer, with the output layer fixed at 49 channels for the 49 eigenvalues (e.g., a first layer with 1024 channels yields 512, 256, 128, and 49 channels). All models were trained for 1000 epochs using the RPD loss and the ADAM optimizer with a learning rate of $10^{-4}$, weight decay of $10^{-5}$, and batch size of 16. Training times ranged from 44 hours for the smallest model to 48 hours for the largest on an NVIDIA A100 GPU (40 GB memory) paired with an Intel Xeon Gold 6248R CPU (64 GB memory), indicating that increased MLP layer width minimally impacts training time. Table \ref{tab:performance_of_various_mlp_layer_width}  reports the training losses and performance metrics.
\begin{table}[ht]
    \centering
    \caption{Testing losses and model performance comparisons with different numbers of hidden channels in the MLP layers. All GCNs were trained with the RPD loss. The number in the right column indicates the number of hidden channels in the first MLP layer. Bold numbers are the best values.}
\label{tab:performance_of_various_mlp_layer_width}
    \begin{tabular}{|l|c|c|c|c|c|}
        \hline
        \textbf{Model} & $\mathbf{L_1}$ & $\mathbf{L_2}$ & \textbf{Polar} & \textbf{RPD} & \textbf{\mbox{PSNR} $>$ 40} \\
        \hline
        $\text{GCN}({1024})$ & 284.82 & 57.35 & 0.0106 & 0.4984 & 0.2764  \\
        \hline
        $\text{GCN}({2048})$ & 252.15 & 51.04 & 0.0078 & 0.3432 & 0.6743 \\
        \hline
        $\text{GCN}({4096})$ & 204.17 & 41.31 & 0.0068 & 0.3058 & 0.7956 \\
        \hline
        $\text{GCN}({8192})$ & \textbf{196.99} & \textbf{39.60} & \textbf{0.0059} & \textbf{0.2705} & \textbf{0.8745}  \\
        \hline
    \end{tabular}
\end{table}

\begin{figure}[H]
    \centering
    \includegraphics[scale=0.375]{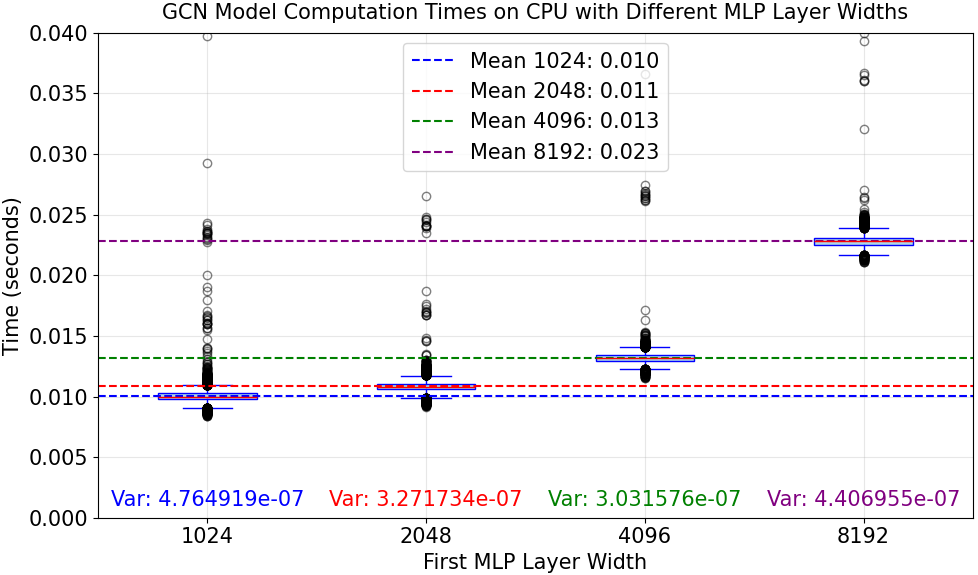}  
    \caption{Prediction times on a CPU for each part in the testing dataset (16,421 CAD meshes) using a GCN model with different first MLP layer widths (1024, 2048, 4096, and 8192), represented by blue, red, green, and purple, respectively. Box plots illustrate the mean and variance of computation times, with variance values annotated. Horizontal dashed lines indicate the mean computation times for each MLP first layer width.}
    \label{fig:Inference_time_comparison_for_different_mlp_layer_width}
\end{figure}

Consistent with the findings in \cite{lee2019wide} and \cite{belkin2021fit}, wider MLP layers (8192 channels in the first layer) enhance convergence speed and prediction accuracy. However, Figure \ref{fig:Inference_time_comparison_for_different_mlp_layer_width} shows that prediction times, particularly on CPU, increase significantly with wider layers, making 8192 channels the most complex architecture considered in this study. The figure illustrates the computation times for four different MLP layer widths—1024, 2048, 4096, and 8192—represented by distinct colors: blue, red, green, and purple, respectively. Each data point is accompanied by a box plot, indicating the mean computation time and its variance, with the variance values labeled below each corresponding width. The results show that as the MLP layer width increases from 1024 to 8192, the mean computation time exhibits a slight upward trend, ranging from approximately 0.010 seconds to 0.023 seconds. However, the variance in computation times remains relatively stable across all widths, with values on the order of $10^{-7}$, suggesting consistent performance stability despite the increasing layer size. This analysis provides insights into the scalability of GCN models on CPU architectures, highlighting the trade-off between layer width and computational efficiency.
\\

\begin{figure}[ht]
    \centering
    \includegraphics[scale=0.325]{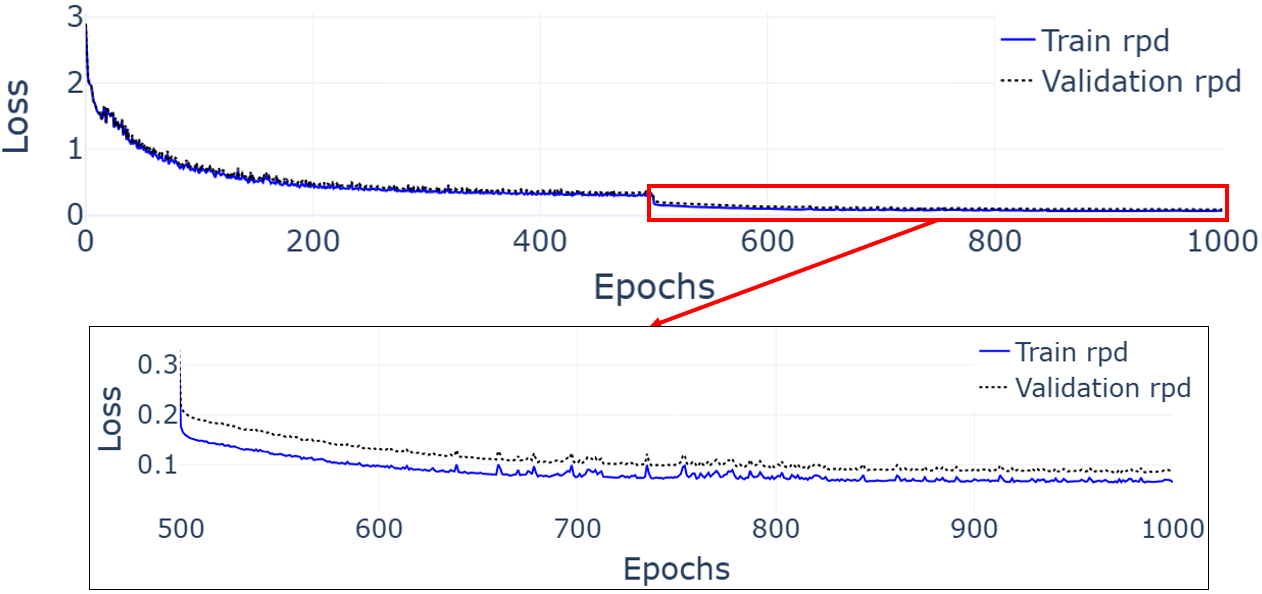}  
    \caption{RPD losses for training data and validation data. The model was first trained for 500 epochs using the ADAM optimizer with $10^{-4}$ as the learning rate, $10^{-5}$ as the weight decay, and 16 as the batch size, then training continued for an additional 500 epochs using a reduced learning rate, $10^{-5}$, while all other parameters remained the same as before.}
    \label{fig:rpd_loss}
\end{figure}

{\em Determining the learning rate schedule.-}
We further optimized the best GCN architecture with wide MLP layers by adopting a reduced learning rate of $10^{-5}$ for epochs 501 to 1000, following an initial learning rate of $10^{-4}$ for the first 500 epochs. Table \ref{tab:performance_constant_vs_reduced_lr} demonstrates that this reduced learning rate strategy significantly outperforms a fixed learning rate of $10^{-4}$ across all 1000 epochs. 
Figure \ref{fig:rpd_loss} displays the training and validation loss of the best GCN model using the RPD loss function under this training regime.
\begin{table}[ht]
    \centering
    \caption{Testing losses and model performance comparisons of best GCN architecture trained with $10^{-4}$ constant learning rate for 1000 epochs against the same model trained with $10^{-4}$ learning rate for 500 epochs and then a reduced learning rate, $10^{-5}$, for the remaining 500 epochs.}
    \label{tab:performance_constant_vs_reduced_lr}
    \begin{tabular}{|l|c|c|c|c|c|}
        \hline
        \textbf{lr} & $\mathbf{L_1}$ & $\mathbf{L_2}$ & \textbf{Polar} & \textbf{RPD} & \textbf{\mbox{PSNR} $>$ 40} \\
        \hline
        \text{constant} & 196.99 & 39.60 & 0.0059 & 0.2705 & 0.8745  \\
        \hline
        \text{reduced} & \textbf{59.84} & \textbf{12.91} & \textbf{0.0018} & \textbf{0.0849} & \textbf{0.9928}  \\
        \hline
    \end{tabular}
\end{table}

\begin{figure}[ht]
    \centering
    \includegraphics[scale=0.5]{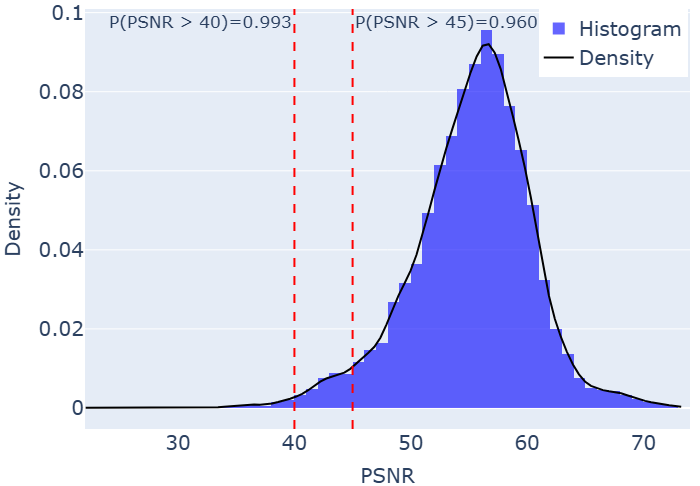}  
    \caption{Distribution of PSNR values for predictions on 16,421 testing samples using the final, best GCN model. $99.3\%$ of the predictions have a PSNR value greater than 40, and $96\%$ of the predictions have a PSNR value greater than 45.}
    \label{fig:psnr_histogram}
\end{figure}

Using the $\mbox{PSNR} > 40$ evaluation metric, the best model achieves 99.3\% accuracy on a test set of 16,421 samples, maintaining 96\% accuracy at a stricter threshold of $\mbox{PSNR} > 45$. Figure~\ref{fig:psnr_histogram} displays the PSNR distribution across all test samples.
\\

\begin{figure}[H]
    \centering
    \includegraphics[scale=0.75]{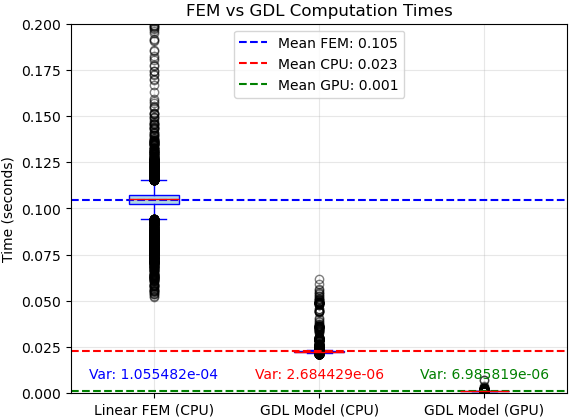}  
    \caption{Computation time comparison of Linear FEM and GCN model predictions (CPU and GPU) for the first 50 eigenvalues across all testing samples.}
    \label{fig:time_comparison}
\end{figure}

{\em Prediction time comparison of best GCN model and linear FEM method.-}
Predicting a mesh with the best found GCN (trained with RPD loss and the varying learning rate scheduled discussed before, with the 8192-wide first MLP layer, as shown in figure \ref{fig:GDL_Architecture}) significantly outperforms the linear FEM method with the Arnoldi algorithm—the fastest FEM approach for LB eigenvalue computation—on both CPU and GPU. Figure~\ref{fig:time_comparison} displays computational times for each of the parts in the testing dataset using: a) the linear FEM on a CPU, b) our best GCN on a CPU, and c) our best GCN on GPU. The linear FEM method, implemented via the LaPy Python package, averages 0.105 seconds per mesh (variance $1.055482 \times 10^{-4}$) on an Intel Xeon Gold 6248R CPU (64 GB memory), while the GCN model averages 0.023 seconds (variance $2.684429 \times 10^{-6}$) on the same CPU and 1 millisecond (variance $6.985819 \times 10^{-6}$) on an NVIDIA A100 GPU (40 GB memory) for predicting the 2nd to 50th LB eigenvalues. Thus, with moderate computational resources, the trained GCN achieves an average 5 times reduction in prediction time relative to FEM (and two orders of magnitude reduction on average if ran on a GPU instead).
\\

\section{Conclusions}
\label{sec:conclusion}
\noindent We have demonstrated that the Laplace-Beltrami spectrum obtained from a mesh CAD model is learnable by a geometric deep learning approach, with predictions comparable to that of linear FEM but at a small fraction of the computing time. We have discussed the final graph convolutional network architecture, which was trained on a specially curated database of real mechanical parts, and we have made available all of our code and datasets. Our approach focused on predicting the 2nd to 50th smallest LB eigenvalues, providing a tool that can be used for predicting the LB spectrum for fairly complex parts (see Figure \ref{fig:Sample_Output} for a few samples, and the complete dataset of parts we are making available--see the Appendix.). 

Some areas for further research remain, based on limitations we now identify. If parts with and even more intricate geometry need to be identified via their LB spectrum (e.g., lattice structures with very complex topology, as produced, for example, with additive manufacturing), the proposed GCN architecture could be scaled up to predict even higher LB eigenvalues, which will capture even finer geometrical details. We leave this as a task for further research. Likewise, we limited the training of the GCN to single-component parts with no boundary and genus less than 3. We speculate that meshes with a boundary can be dealt with via training, given that the FEM laplacian can handle such cases. Also, higher genus parts could be handled, provided more eigenvalues are learned and predicted instead with straightforward modification of the MLP layers' widths (how many more eigenvalues depends on the specific part, see \cite{an2025practical}). This will inevitably require more computing time for generating predictions, but this should be a feasible alternative over linear FEM, given that the computing time to obtain the LB spectrum via linear FEM also increases with the number of eigenvalues desired. Therefore, future research should focus on modifying the GCN architecture presented here to predict perhaps up to a few hundred eigenvalues to handle more complex topologies, comparing against the linear FEM computing time, thereby generalizing the present work for application to even more complex CAD models. Therefore, the present work establishes a scalable foundation for operator learning, advancing geometric learning for real-world engineering applications.

\section*{Data Availability Statement.}
\label{app: supplementary}
\noindent The supplementary materials contain the curated mechanical CAD models with 32841 parts derived from the ABC benchmark dataset, the best GCN model, and all Python scripts, including performing data pre-processing, extracting geometric features, constructing the GCN models, and training and testing as presented in the paper, with instructions on how to use. The supplementary materials are available to download at: \\
\url{https://drive.google.com/drive/folders/1LAPl_khJ3VcO1YldGez6p7XWcDCiOK5S}.

\section*{Disclosure of interest.} The authors have no conflicts of interest to report.

\section*{Acknowledgments.} This work was funded by NSF grant CMMI 2121625.

\bibliographystyle{plain}

\end{document}